# Emory Knee Radiograph Dataset

Brandon Price[1], Jason Adleberg[2], Kaesha Thomas[3], Zach Zaiman[4], Aawez Mansuri[5], Beatrice Brown-Mulry[4], Chima Okecheukwu[6], Judy Gichoya[3], Hari Trivedi[3]

**Affiliations**
1. Department of Radiology, University of Florida, Gainesville, FL
2. Department of Radiology, Mount Sinai Health System, New York City, NY
3. Department of Radiology, Emory University, Atlanta, GA
4. Department of Computer Science, Emory University, Atlanta, GA
5. School of Medicine, Emory University, Atlanta, GA
6. Department of Computer Science, Georgia Institute of Technology, Atlanta, GA

Corresponding author: Brandon Price (bpri0002@radiology.ufl.edu, bprice9@emory.edu)

## Abstract

The Emory Knee Radiograph (MRKR) dataset is a large, demographically diverse collection of 503,261 knee radiographs from 83,011 patients, 40% of which are African American. This dataset provides imaging data in DICOM format along with detailed clinical information, including patient-reported pain scores, diagnostic codes, and procedural codes, which are not commonly available in similar datasets. The MRKR dataset also features imaging metadata such as image laterality, view type, and presence of hardware, enhancing its value for research and model development. MRKR addresses significant gaps in existing datasets by offering a more representative sample for studying osteoarthritis and related outcomes, particularly among minority populations, thereby providing a valuable resource for clinicians and researchers.

# Introduction

Millions of Americans suffer from knee pain, with the predominant cause being osteoarthritis, leading to decreased mobility and quality of life [1,2,3]. Previous research has shown that estimating patient pain based on knee radiographs is suboptimal, particularly for minority patients [4,5]. Deep learning models can be trained to use knee radiographs to grade osteoarthritis severity and predict pain scores for patients, helping clinicians and researchers gain deeper insights into osteoarthritis [6,7,8,9,10,11]. However, there are limited publicly available datasets for knee radiographs, and none are both demographically diverse and include patient outcomes and pain scores. This can lead to poor generalizability of models trained using these data, particularly for minority patients.[12,13,14]

While several robust longitudinal knee radiograph datasets exist, they are generally small for deep learning model development, and none include patient CPT (Current Procedural Terminology) and ICD (International Classification of Diseases) codes. For example, the Osteoarthritis Initiative an extensive multicenter dataset of biomarkers, survey responses, and knee radiographs of 4,796 patients over nine years[15]. Similarly, the Multicenter

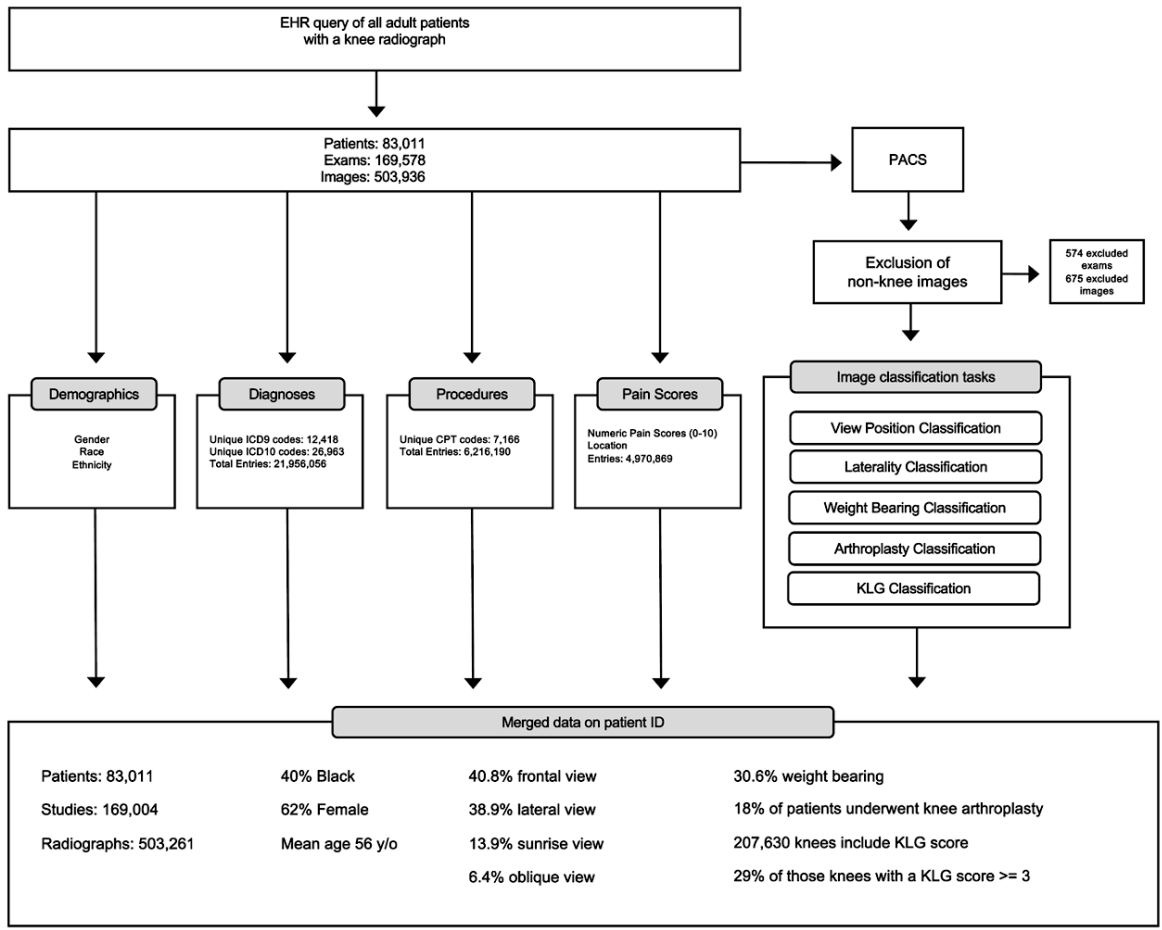

Figure 1: Workflow for selecting and classifying knee radiographs and associated clinical data. The initial query yielded 83,011 patients, yielding 169,578 exams and 503,936 images. Clinical data included demographics (gender, race, ethnicity), diagnoses categorized by unique ICD-9 and ICD-10 codes, procedures classified by CPT codes, and numerical pain scores and their specific locations. Inference from models trained to classify radiograph attributes such as view position, laterality, weight-bearing status, arthroplasty detection, and KLG scoring were included. The data streams were merged on patient ID, providing insights into patient demographics, examination details, and clinical outcomes.

Osteoarthritis Study (MOST), consists of 20,067 knee radiographs collected from 4,446 patients over seven years, providing prospective observational data[16]. The Lower Extremity Radiographs (LERA) dataset, released by Stanford University, also [17] knee radiographs, along with foot, ankle, and hip radiographs . This dataset includes 13,106 patients and labels indicating normal or abnormal findings based on a radiology report. FracAtlas, a dataset from Bangladesh, 4,083 radiographs annotated with fracture types, localizations, and segmentations[18]. Of these radiographs, 2,272 are of the lower extremities. Although the NIH and MOST datasets are comprehensive and include numerous biomarkers with patient survey responses, they remain limited in representing Black Americans and the quantity of patients overall.

Our Emory Musculoskeletal Knee Radiograph Dataset (MRKR) contains 503,261 knee radiographs of 83,011 patients, 33,503 (40.4%) of whom are African American. The dataset includes patient pain scores, ICD codes, and CPT codes for knee-related procedures and outcomes. Each image is also curated image laterality, view type, Kellgren-Lawrence osteoarthritis severity grading score (KLG) [19,20], and hardware annotations. This makes MRKR amongst the largest and most demographically diverse dataset of its kind.

## Material and Methods

The Institutional Review Board approved the development of this retrospective dataset, comprising knee radiographs and clinical data. The data was compiled through a combination of primarily automated and semi-automated curation techniques, further detailed below. Figure 1 provides an overview of the full dataset.

## Data Extraction

### *Patient Identification*
We identified 83,011 adult patients (age >= 18) who received knee radiographs between 2002 and 2021 from four affiliated hospitals, which include two community hospitals, one urban hospital, and one academic hospital. This included outpatient, inpatient, emergency department, and pre and post-surgical radiographs, including those obtained by orthopedic surgeons, which were not interpreted by radiologists.

### *Imaging data*
Exams were extracted from the institution's Picture Archiving and Communications System (PACS) in DICOM format using Niffler[21], an open-source pipeline developed in-house for retrospective image extraction that utilizes pydicom[22].

DICOM metadata was de-identified using an internal pipeline to maintain relationships and temporality between exams from the same patient and accompanying clinical data. A pixel de-identification tool from MD.ai (www.md.ai) was used to remove any pixel data, such as names, dates, or times, from the DICOM images. DICOMs that included non-radiographic modalities or secondary captures were dropped.

### *Clinical Coding Data*
ICD and CPT codes were extracted from the Clinical Data Warehouse (CDW), which regularly receives information from the Electronic Health Record. ICD codes were in ICD 9 format prior to 2016 and ICD 10 afterwards, and are recorded as such. CPT codes included two character modifiers often used to indicate the anatomic location and laterality of a procedure. A de-identified encounter number is included in order to link ICD and CPT data. Patient ID and date information were anonymized using the same method used for imaging.

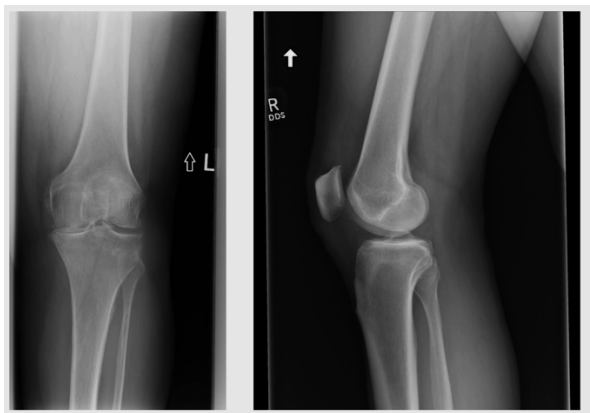

Figure 2: (Right) Unilateral left knee radiograph in the frontal projection which is weight bearing as indicated by the upward pointing arrow. (Left) Lateral view of a right knee which is weight bearing.

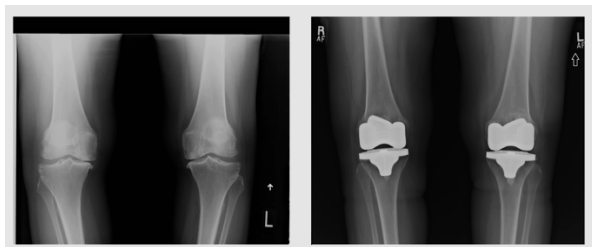

Figure 3: (Left) Bilateral weight bearing knee radiograph in the frontal projection. There is moderate medial compartmental osteoarthritis bilaterally. (Right) Same patient status post bilateral knee arthroplasty.

| Dataset Patient Summary Statistics | | |
|---|---|---|
| Total Patients | 83,011 | 100% |
| Gender | | |
|     Female | 51,175 | 61.6% |
|     Male | 31,836 | 38.4% |
| Age, years | | |
|     Mean | 59.2 std. +/- 15.4 | |
|     Median | 61 | |
| Race demographics | | |
|     White | 36,927 | 44.5% |
|     Black | 33,503 | 40.4% |
|     Asian | 2,893 | 3.5% |
|     Unknown/Unreported | 8,751 | 10.5% |
|     Multiple | 536 | 0.6% |
|     American Indian or Alaskan Native | 244 | 0.3% |
|     Native Hawaiian or Other Pacific Islander | 157 | 0.2% |
| Ethnicity | | |
|     Hispanic | 2,501 | 3.0% |
|     Non-Hispanic | 66,378 | 80.0% |
|     Unknown/Unreported | 14,132 | 17.0% |
| Clinical outcomes | | |
|     Arthroplasty | 14,843 | 17.9% |

*Table 1: Dataset Patient Summary Statistics*

*Patient-reported pain data*

Pain scores and associated dates were extracted from the CDW. Pain data is recorded as an integer score from 0 to 10 with a free-text location of pain. Parsing free-text to categorize pain location was challenging as the location could be mentioned as an extremity (i.e. left lower extremity), a joint (i.e. knee), or more vaguely (i.e. leg). First, a series of regular expression rules were used to filter definitive knee-related pain scores in the dataset. In cases where the pain location was indicated as "lower extremity" or other vague description, the pain location value was assigned "other" due to ambiguity.

*Demographic data*

Patient self-reported race, gender, and ethnicity are extracted at the patient level. Patient age at the time of exam is abstracted using de-identified data of birth information. Race categories for patients consist of White, Black, Asian, American Indian or Alaskan Native, Native Hawaiian Other Pacific Islander, or multiple. A small fraction, less than one percent, were identified as multiple races. For ethnicity, responses were distilled into three groups: Hispanic, non-Hispanic, and unknown/unreported. This categorization is based the varying syntax of responses.

## Data Curation

We discovered several inconsistencies in the original imaging data, including exam descriptions for knee radiographs that contained other erroneous anatomic regions such as chest or hand radiographs, or did not include the entire knee joint. We trained a ConvNeXt model to exclude non-knee images by labeling 8,362 knee and non-knee images using the MD.ai annotation platform. Overall accuracy for detecting non-knee radiographs was 0.999, with an F1-score of 0.961 with a 5% prevalence. Using this model, we discovered 675 non-knee radiographs which were excluded from the dataset.

We found DICOM metadata unreliable for categorizing image laterality and view type in radiographs, often mislabeling unilateral as bilateral images and vice versa, and frequently missing or inaccurately noting view positions like frontal, lateral, oblique, and sunrise. See figures 2 and 3 for examples of knee radiograph view positions and laterality. There was no reliable metadata for weight bearing studies or presence arthroplasty (total, hemi, or unicompartmental). To address this, we manually labeled 6,000 radiographs for these details using MD.ai and trained a multi-class ConvNeXt classifier. The model achieved high weighted average F1-scores across categories: 0.985 for laterality, 0.974 for view position, 0.981 for weight bearing, and 0.992 for arthroplasty. When images lacked a laterality marker (n=2647), we used the DICOM metadata for image laterality.

Finally, a state-of-the-art, open-source model developed by Duke University was used to predict the Kellgren Lawrence Grading Scale (KLG)[23] score per knee which measure osteoarthritis severity. The KLG score ranges from 0 (no radiographic evidence of osteoarthritis) to 4 (severe features of osteoarthritis). In prior work. this model was 75.9% accurate at predicting KLG scores using the MOST dataset[24] and of the 24.1% of knees that were incorrectly classified, 93.5% were off by one grade. The model only provides KLG inference for bilateral frontal view knee radiographs without knee arthroplasty. No systematic manual review or verification of model accuracy was done on MRKR, however an anecdotal review of several hundred images showed good performance, suggesting the model generalized to our dataset.

# Resulting Dataset

## *Patient Data Characteristics*

Patient information is summarized in Table 1. This dataset includes 83,011 patients, 61.7% of whom are women. The mean patient age was 59.2 (std +/- 15.4 years). Self-reported racial distribution was 44.5% White, 40.4% Black, 3.5% Asian, less than 1% for American Indian, Native Hawaiian, and multiple. The average age for White patients was 62.0 (std +/- 15.1) and the average for Black patients was 57.6 (std +/- 14.6). For approximately 10.5% of patients the racial data is either unknown or was not reported. Self-reported ethnicity was 3.0% Hispanic, 80.0% non-Hispanic, and 17.0% unknown or unreported.

## *Imaging Data Characteristics*

This dataset includes (503,261) knee radiographs from 169,004 exams. Each patient has an average of 2.0 studies per patient (std +/- 2.1) and a mean of 6.1 knee radiographs (std +/- 5.9). The mean KLG score was 1.8 (std +/- 1.3). Prevalence of arthroplasty or arthrodesis was 17.88% at the patient level and 22.8% at the exam level.

## *Clinical Data Characteristics*

There was a mean of 264.5 ICD codes and 74.9 CPT codes per patient. Some of the ICD codes, such as essential hypertension, were frequently repeated within a patient across different dates, which are included in the dataset. A total of 4,970,869 pain scores were recorded across all patients, with a mean of 59.9 (std +/- 160.3) pain scores per patient. The mean and median reported knee pain scores were 4.2 (std +/- 3.4) and 4, respectively. Average reported knee pain for White patients was 3.6 (std +/- 3.2), while average reported knee pain for Black patients was 4.7 (std +/- 3.5). Figures 4 and 5 further characterize the differences in knee pain among racial subgroups, considering KLG scores and before and after arthroplasty.

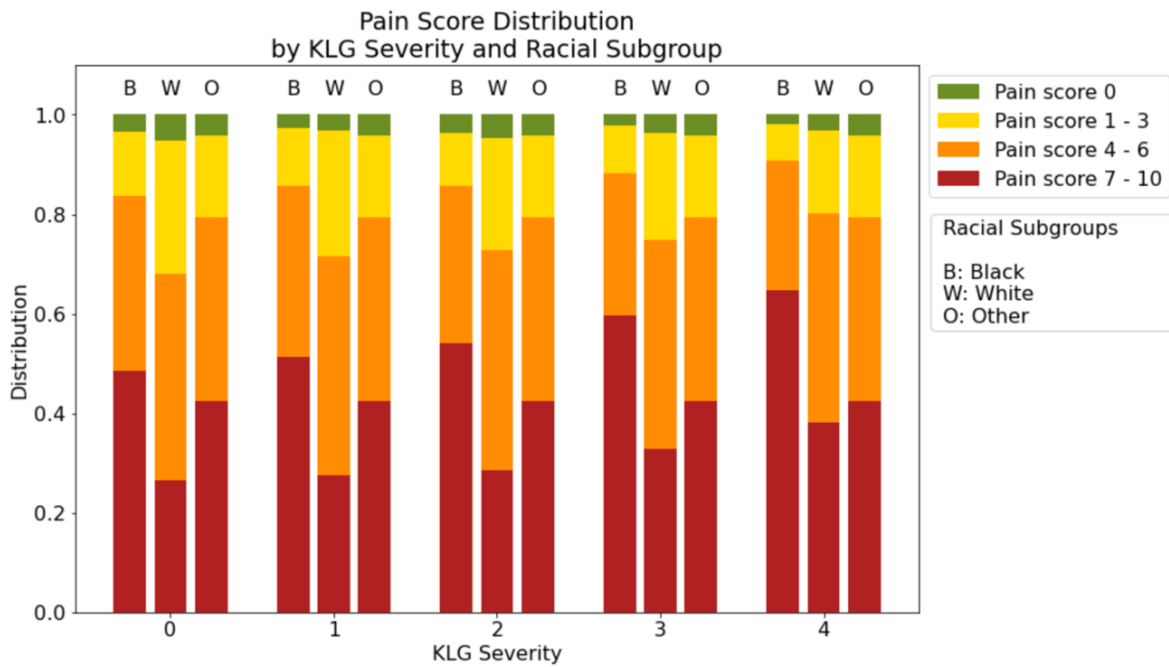

Figure 4: Distribution of pain scores across KLG severity levels by racial subgroup (Black, White, and Other).

*Database size and structure*

The total dataset is 2.3 TB and includes DICOMs (2.3 TB) and seven CSV files (2.7 GB), each containing clinical and metadata.

## Discussion

The Emory Musculoskeletal Knee Radiograph (MRKR) dataset, includes 503,261 knee radiographs from 83,011 racially and ethnically diverse patients at a single institution. This dataset is enhanced by extensive clinical data, including pain scores, ICD, and CPT codes. Images are also annotated with information not traditionally available or unreliable in DICOM metadata including presence of hardware, views, laterality, and weightbearing status.

The MRKR dataset is not without limitations. The Duke University KLG prediction model used for frontal radiograph KLG inference was not systematically verified with this specific dataset, possibly affecting the precision of osteoarthritis severity assessments. Additionally, the subjectivity inherent in patient-reported numeric pain scores, influenced by psychosocial factors and cultural backgrounds, presents challenges in inter-individual variability. These pain scores do not clearly distinguish between constant and intermittent pain or accurately reflect intensity fluctuations. While more detailed pain reporting scales like KOOS or WOMAC exist, they are not commonly used in clinical or hospital settings. Lastly, pain scores pertinent to the knee may not be accurately recorded if they were documented as 'lower extremity' which means some patient pain scores may not be represented.

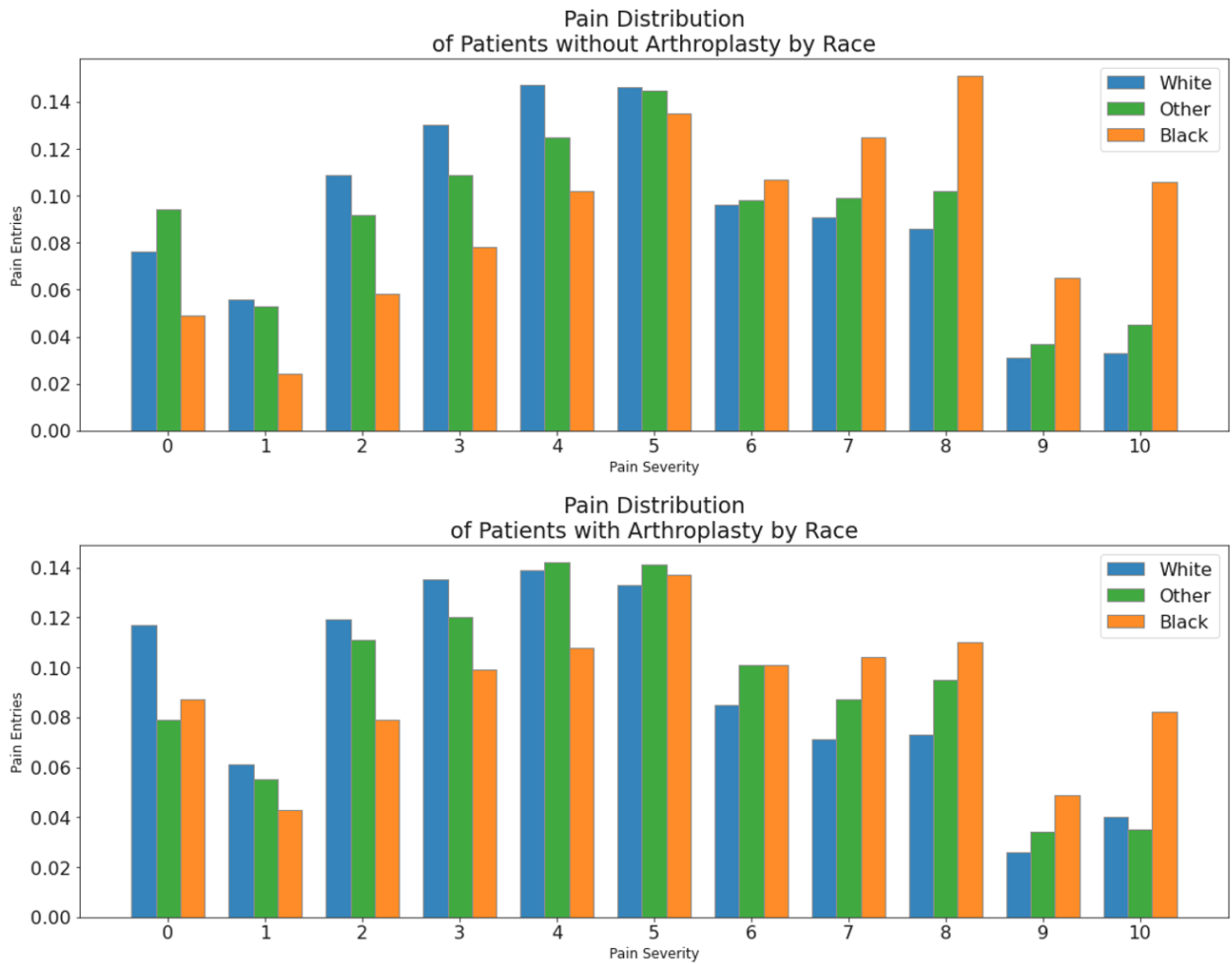

Figure 5: Distribution of pain scores by patient racial subgroup (White, Black and other) before and after arthroplasty.

In summary, the MRKR dataset is a comprehensive collection of over half a million knee radiographs coupled with extensive clinical data. The dataset's diversity, particularly its inclusion of a significant proportion of African American patients, addresses a critical gap in existing datasets. The dataset's inclusion of patient-reported pain scores alongside clinical diagnoses and procedures offers an opportunity to correlate radiographic findings with subjective pain experiences and outcomes, providing a deeper understanding of osteoarthritis and its impact on patients.

# Citations

# Dataset Structure

Filename: MRKR_CPT.csv
File size: 178 MB
Total rows: 6,216,190

Description: This table contains information regarding all CPT codes for a patient and corresponding dates

| Field Name | Data Type | Description |
| --- | --- | --- |
| empi_anon | Integer (8 digits) | Unique patient identification number (83,011 patients) |
| cpt_code | String 5 characters | Current Procedural Terminology code used in coding of medical services and procedures for billing (7,166 CPT codes) |
| cpt_group_modifier | String | Used to provide further information regarding service or procedure. Most CPT codes do not include modifier data. If there is modifier data, it is often used to indicate the laterality of a procedure (left or right). There can be multiple modifiers for a single CPT code entry. This field combines all of the modifiers for that particular CPT code for that specific encounter for that patient. |
| date_anon | Date | Date of when the associated procedure or service occurred. |
| age_at_procedure | Integer | Age when the procedure was performed |

File name: MRKR_CPT_dictionary.csv
Filesize: 754 KB
Total rows: 7,166

Description: A lookup table between CPT codes and corresponding descriptions

| Field Name | Data Type | Description |
| --- | --- | --- |
| cpt_code | String 5 characters | Current Procedural Terminology code used in the coding of medical services and procedures for billing |

| cpt_description | String | Description of the procedure. There are some unique CPT codes that share the same description. This leads to 128 fewer unique descriptions compared to unique CPT codes. |

Filename: MRKR_ICD.csv
File save: 1.7 GB
Total rows: 21,956,056

Description: ICD9 and ICD10 diagnoses for patients with corresponding dates. Certain diseases of interest are indicated by binary flags to ease data cleaning.

| Field Name | Data Type | Description |
| --- | --- | --- |
| empi_anon | Integer (8 digits) | Unique patient identification number<br>83,011 unique patients |
| ICD9 | String | International Classification of Diseases - 9<br>12,418 unque codes |
| ICD10 | String | International Classification of Diseases - 10<br>26,963 unique codes |
| date_anon | Date | Date of when the diagnosis code was entered. |
| age_at_dx | Integer | Age when the diagnosis was recorded |
| DX_LINE | String | Primary, Secondary, Active, Not Recorded, Resolved, Canceled, Inactive |
| DX_ICD_SCOPE | String | Billing Diagnosis, Discharge Diagnosis, Admitting Diagnosis, Referring Diagnosis, Not Recorded, Reason For Visit, Problem List, Working Diagnosis, Other Diagnosis, Final, Pre-Op Diagnosis, Post-Op Diagnosis, Principal Diagnosis, Suggested Billing |
| autoimmune | Binary | If ICD code corresponds to auto-immune disease such as rheumatoid arthritis, juvenile arthritis, gout, etc.<br><br>ICD10 prefixes; M05, M06, M08, M10, M45 & M1A<br>ICD9 prefixes; 274, 714, & 720<br>122,859 entries<br>9,704 patients (11.7%) |
| diabetes | Binary | If ICD code corresponds to type I or type II diabetes<br><br>ICD10 prefixes, E08, E10, E11 & E13<br>ICD9 prefix 250<br>552,208 entries<br>18,655 patients (22.47%) |
| hypertension | Binary | If ICD code corresponds to hypertension<br><br>ICD10 prefixes I10-I16<br>ICD9 prefixes 401-405<br>1,209,572 entries<br>45,300 patients (54.57%) |
| joint_infection | Binary | If ICD code corresponds to a knee joint infection<br><br>ICD10 prefixes 'M00.06', 'M00.16', 'M00.26', 'M00.86', 'M01.X6', 'M02.06', 'M02.16', 'M02.26', 'M02.36', 'M02.86'<br>ICD9 prefixes '711.06', '711.16', '711.26', '711.36', '711.46', '711.56', '711.66', '711.76', '711.86', '711.96'<br>2,224 entries<br>551 patients (0.66%) |
| knee_osteoarthritis | Binary | If ICD code corresponds to knee osteoarthritis<br><br>ICD10 prefixes M17<br>ICD9 prefixes 715.16, 715.26, 715.36, 715.96<br>373,186 entries<br>51,468 patients (62.0%) |
| knee_osteomyelitis | Binary | If ICD code corresponds to knee osteomyelitis<br><br>ICD10 prefixes M86 related to thigh and tibia/fibula.<br>ICD9 prefixes 730 related to lower leg<br>7,130 entries<br>1,389 patients (1.7%) |
| obesity | Binary | If ICD code corresponds to knee obesity<br><br>ICD10 prefix E66.01, E66.09, E66.1, E66.3, E66.8, E66.9<br>ICD9 prefix 278.0, 278.00, 278.01, 278.02 |

| | | 197,094 entries<br>27,576 patients (33.2%) |
|---|---|---|
| nicotine_use | Binary | If ICD code corresponds to nicotine dependence<br><br>ICD 10 prefix F17, Z57.31, Z71.6, Z72.0, Z77.22, Z87.891<br>ICD9 305.1<br>148,095 entries<br>20,882 patients (25.2%) |
| trauma_lower_extremity | Binary | If ICD code corresponds to lower extremity trauma<br><br>ICD10 prefixes S82.0, S82.1, S83<br>ICD9 prefixes 959.7, 844, 823, 891, 822, 836<br>152,377 entries<br>34,230 patients (41.2%) |

Filename: MRKR_ICD_dictionary.csv
File size: 1.9 MB
Total rows: 25,209

Description: Lookup table for ICD9 and ICD10 codes and corresponding descriptions.

| Field Name | Data Type | Description |
|---|---|---|
| ICD9 | String | ICD9 code |
| ICD10 | String | ICD10 code |
| DX_NAME | String | Diagnosis name or description |

Filename: MRKR_pain.csv
File size: 137 MB
Total rows: 4,975,933

Description: Contains information on self-reported pain scores by patients during any encounter, including outpatient, emergency, and perioperative. Pain scores related to knees are curated.

| Field Name | Data Type | Description |
|---|---|---|
| empi_anon | Integer (8 digits) | Unique patient identification number 83,011 unique patients |
| pain_location | String | Raw, uncurated strings of pain locations entered by staff. Approximately 75% of entries are blank. |
| knee_pain | Binary | Curated using regular expressions to identify if the pain_location is definitely knee related. |
| pain_score | Integer | 0 - 10 pain score |
| laterality | Nominal string | [R: Right, L: Left, B: Bilateral, NaN: Unknown or not present] Value only present if knee_pain is 1 |
| date_anon | Date | Date of when the pain score was entered into the patient's chart. |

Filename: MRKR_demographics.csv
File size: 4.5 MB
Total rows: 83,011

Description: Patient demographics, indexed at the patient level.

| Field Name | Data Type | Description |
|---|---|---|
| empi_anon | Integer (8 digits) | Unique patient identification number |
| sex | Nominal string | [male, female] Patient sex |
| race | Nominal string | [African American or Black, American Indian or Alaskan Native, Asian, Caucasian or White, Multiple, Native Hawaiian or Other Pacific Islander, |

|  |  | Unknown] |
|  |  | Patient self-reported race |
| ethnicity | Nominal string | [Hispanic patients,<br>Non-Hispanic patients,<br>Unknown,<br>Unreported]<br><br>Patient reported ethnicity |

Filename: MRKR_image_metadata.csv
File size: 210 MB
Total rows: 503,261

Description: Contains relevant public DICOM metadata tags that may be helpful for identifying images. Patient and exam identifiers are replaced with de-identified versions in this table and within DICOM files. Other Non-PHI containing metadata tags that are not in this table are left intact within DICOM files. Fields containing PHI such as patient name, addresses, or referring physician are removed from this table and DICOM files. For data curation, the below fields were modified or added.

| Field Name | Data Type | Description |
| --- | --- | --- |
| empi_anon | Integer (8 digits) | De-identified patient identification number |
| StudyInstanceUID_anon | String | De-identified Study UID, shared between all images in the same study |
| SeriesInstanceUID_anon | String | De-identified Series UID, shared between all images in the same series |
| SOPInstanceUID_anon | String | De-identified SOP Instance UID which corresponds to a single DICOM image |
| img_height | Integer | Image pixel height |
| img_width | Integer | Image pixel width |
| laterality | Nominal string | [R: Right,<br>L: Left,<br>B: Bilateral, |

| | | -1: Unknown or not present] |
| | | Laterality of the image, as inferred by DL model |
| view_position | Nominal string | [F: Frontal,<br>L: Lateral,<br>S: Sunrise,<br>I: Internal Oblique,<br>E: External Oblique]<br><br>Anatomical projection of radiograph, as inferred by DL model |
| horizontal_flip | Binary | Indicates if the patient's left side was oriented to the left side of the image, which is opposite of typical radiographic orientation, as inferred by DL model. |
| weight_bearing | Binary | Indicates if the radiograph was weight-bearing as indicated by a marker and derived by DL model. Not all images in a given exam will be weight-bearing or non-weightbearing. |
| inverted | Binary | Indicates whether pixel intensity values are inverted from typical radiographic convention, as inferred by DL model |
| arthroplasty | Nominal string | [R: right,<br>L: left,<br>B: bilateral,<br>NL: unknown (no laterality marker),<br>NaN: no arthroplasty]<br><br>Indicates if image contains a knee arthroplasty and its laterality, as derived by a DL model. Applies to unilateral and bilateral views. |
| L_KLG_inference | Ordinal integer | [0,1,2,3,4, NaN]<br><br>KLG score of left knee in a bilateral knee radiograph, inferred by DL model |
| R_KLG_inference | Ordinal integer | [0,1,2,3,4, NaN]<br><br>KLG score of right knee in a bilateral knee radiograph, inferred by DL model |

| SeriesDescription | String | DICOM Metadata describing the series |
| StudyDescription | String | DICOM metadata describing the study |
| StudyDate_anon | Date | De-identified date of radiograph |
| age_at_exam | Integer | Age of the patient when the radiograph was performed |
| dicom_path | String | Path to DICOM file |